\def\year{2021}\relax
\documentclass[letterpaper]{article} 
\usepackage{aaai21}  
\usepackage{times}  
\usepackage{helvet} 
\usepackage{courier}  
\usepackage[hyphens]{url}  
\usepackage{graphicx} 
\usepackage{graphicx}  
\usepackage{natbib}  
\usepackage{caption} 
\usepackage{booktabs}
\usepackage{multirow}
\usepackage{subfigure}
\usepackage{color}
\usepackage[flushmargin]{footmisc} 
\usepackage{graphicx}
\usepackage{grffile}
\usepackage{amsmath}
\usepackage[switch]{lineno}
\title{Uncertainty-Aware Multi-View Representation Learning}
\author{
	Yu Geng, \textsuperscript{\rm 1}
	Zongbo Han, \textsuperscript{\rm 1}
	Changqing Zhang,  \textsuperscript{\rm 1,2}\footnote{Corresponding author}
	Qinghua Hu  \textsuperscript{\rm 1,2}
}
\affiliations{
	
	\textsuperscript{\rm 1} College of Intelligence and Computing, Tianjin University, Tianjin, China\\
	\textsuperscript{\rm 2} Tianjin Key Lab of Machine Learning, Tianjin, China
	
	\{gengyu, zhangchangqing, huqinghua\}@tju.edu.cn,~
	hanzb1997@gmail.com
	
}

\begin{document}
	\maketitle
	
	\begin{abstract}
		Learning from different data views by exploring the underlying complementary information among them can endow the representation with stronger expressive ability. However, high-dimensional features tend to contain noise, and furthermore, the quality of data usually varies for different samples (even for different views), i.e., one view may be informative for one sample but not the case for another. Therefore, it is quite challenging to integrate multi-view noisy data under unsupervised setting. Traditional multi-view methods either simply treat each view with equal importance or tune the weights of different views to fixed values, which are insufficient to capture the dynamic noise in multi-view data. In this work, we devise a novel unsupervised multi-view learning approach, termed as Dynamic Uncertainty-Aware Networks (DUA-Nets). Guided by the uncertainty of data estimated from the generation perspective, intrinsic information from multiple views is integrated to obtain noise-free representations. Under the help of uncertainty, DUA-Nets weigh each view of individual sample according to data quality so that the high-quality samples (or views) can be fully exploited while the effects from the noisy samples (or views) will be alleviated. Our model achieves superior performance	in extensive experiments and shows the robustness to noisy data.
	\end{abstract}
	
	\section{Introduction}
	In recent years, there is a growing interest in multi-view learning. Information in the real world is usually in different forms simultaneously. When watching videos, the optic nerve receives visual signal while the auditory nerve receives speech signal. These two different types of signals complete each other and provide more comprehensive information. Accordingly, conducting representation learning from multi-view data has the potential to improve data analysis tasks \cite{yang2018multi,baltruvsaitis2018multimodal,li2018survey}.
	
	However, the relationship among multiple views is usually very complex. There are two well-known principles in multi-view learning, i.e., \textit{consistency} and \textit{complementary} \cite{li2018survey,zhang2020deep}. Most existing methods mainly focus on the consistency of multiple views which assume that the correlations among views should be maximized \cite{kumar2011coR,Wang2016On}. While there is also complementary information that is vital to comprehensive representations. Therefore, some methods are proposed to explore the complete information of multi-view data \cite{zhang2019ae2,hu2019doubly}. More importantly, different sources of data may contain different amounts of information and possible noise. For example, due to the various sensor qualities or environmental factors, the information of different observations varies from each other. The quality of data usually varies for different samples (even for different views), i.e., one view may be informative for one sample but not the same case for another. Above challenges make multi-view learning rather difficult. In the context of unsupervised representation learning, it is even more challenging due to the lack of label guidance.
	
\begin{figure*}[tbp]
		\centering 
		\subfigure[]{
			\includegraphics[width=8cm]{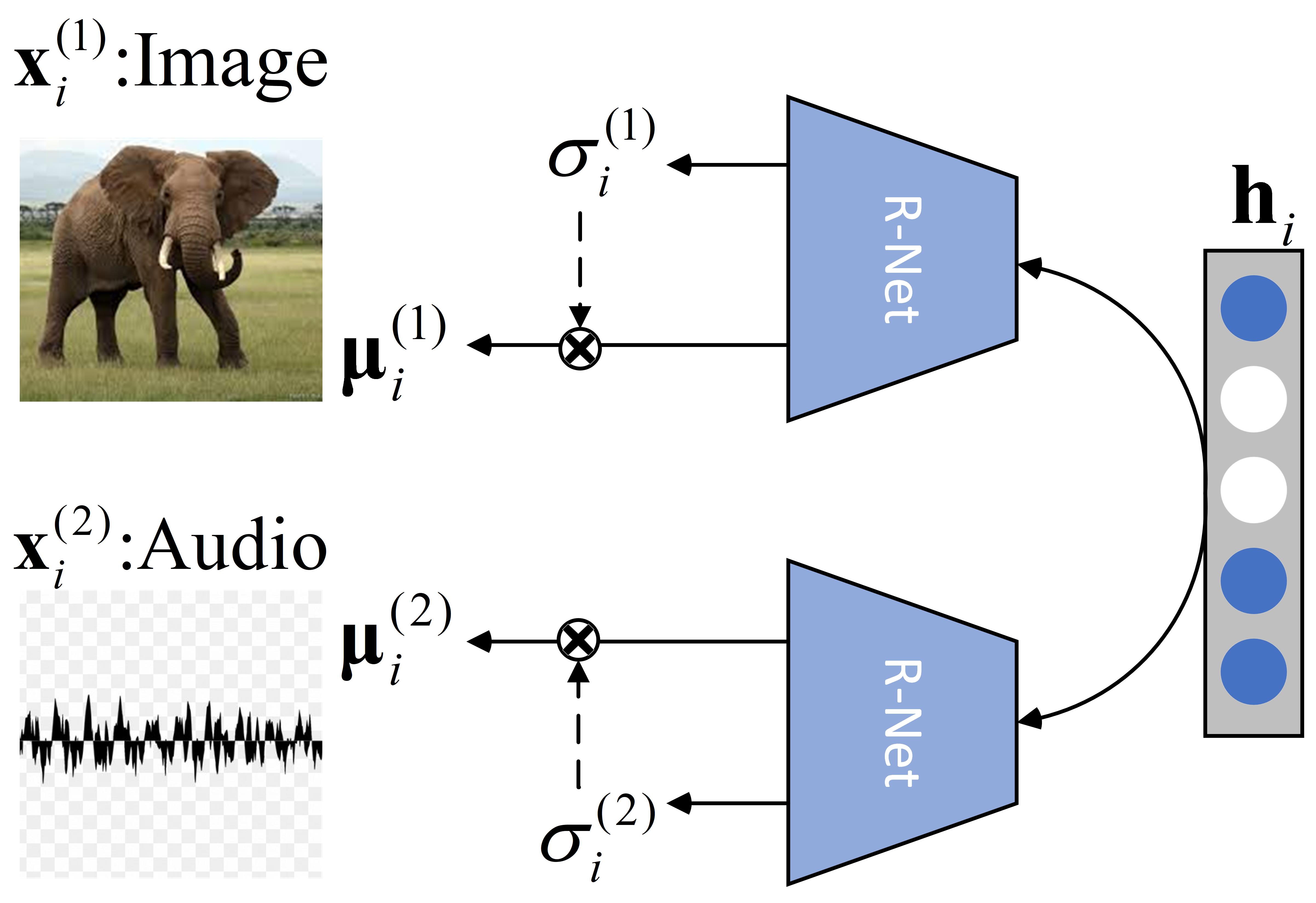}}
		\subfigure[]{
			\includegraphics[width=8cm]{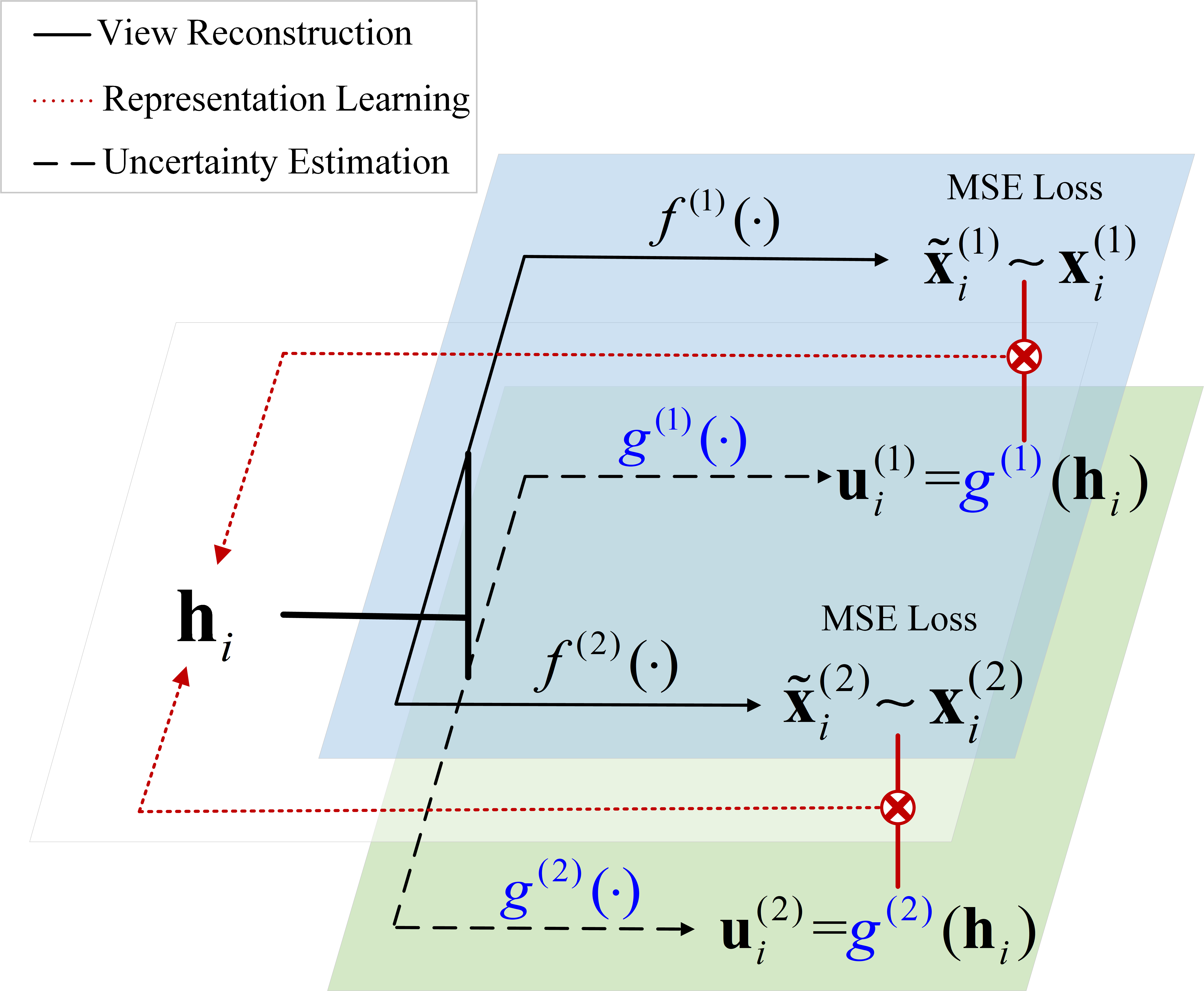}}
		\caption{(a) Overview of the proposed DUA-Nets. (b) Learning process. We use two views for better elaboration. Latent variable $\mathbf{h}_i$ reconstructs each view through $f^{(v)}(\cdot)$. Simultaneously, $g^{(v)}(\cdot)$ estimates the uncertainty in the $v^{th}$ view, which reflects the quality of view reconstruction. The learned uncertainty and reconstruction loss jointly guide the learning of $\mathbf{h}_i$.}
		\label{fig:framework}
	\end{figure*}

	In this work, we propose a novel algorithm termed Dynamic Uncertainty-Aware Networks (DUA-Nets) to address these issues. As shown in Fig. \ref{fig:framework}, we employ Reversal Networks (R-Nets) to integrate intrinsic information from different views into a unified representation. R-Nets reconstruct each view from a latent representation, and thus the latent representation can encode complete information from multiple views. Furthermore, we are devoted to modeling the quality of each sample-specific view. This is quite different from the straightforward ways which ignore the differences between views and samples \cite{andrew2013deep,Wang2016On}. Another common approach is assigning each view a fixed weight \cite{huang2019auto,peng2019comic}. Although it considers view differences and is more effective than equal weighting, it cannot be adaptive to the noise variation inherent in different samples. In this paper, we employ \textit{uncertainty} to estimate the quality of data. Specifically, under the assumption that each observation is sampled from a Gaussian distribution, R-Nets are applied to generate the mean and variance of the distribution, where the variance determines the sharpness of Gaussian distribution, and thus can be interpreted as uncertainty. Modeling data uncertainty can adaptively balance different views for different samples, which results in superior and robust performance. Comprehensive experiments demonstrate the effectiveness of the proposed DUA-Nets. We further provide insightful analyses about the estimated uncertainty.
	
	For clarification, the main contributions of this work are summarized as:
	\begin{enumerate}
		\item We propose an unsupervised multi-view representation learning (UMRL) algorithm which can adaptively address samples with noisy views, and thus, it guarantees the intrinsic information of multiple views are encoded into the learned unified representation.
		\item We propose a novel online evaluation strategy for data quality by using uncertainty modeling, where the uncertainty can guide multi-view integration and alleviate the effect of unbalanced qualities of different views.
		\item We devise a collaborative learning mechanism which seamlessly conducts representation learning and uncertainty estimation in a unified framework so that they can improve each other adaptively.
		\item We conduct extensive experiments to validate the effectiveness of the proposed algorithm. In addition, insightful analyses are provided to further investigate the estimated uncertainty.
	\end{enumerate}
	
	\section{Related Works}
	\paragraph{Multi-View Representation Learning.} The core problem of multi-view learning is how to effectively explore the consistency and complementary information of different views. Plenty of research works focus on multi-view learning and have achieved great progress. The most representative methods are canonical correlation analysis (CCA) \cite{hotelling1936relations} and its variants \cite{bach2002kernel,hardoon2011sparse,andrew2013deep,Wang2016On}. CCA searches a  shared embedding of two views through maximizing the correlation between them. To reduce the influence of noisy data, sparse CCA \cite{hardoon2011sparse} is proposed to learn sparse representations. Kernel CCA \cite{bach2002kernel} extends CCA to nonlinear conditions, which is the case of most real-world multi-view data. Based on deep learning framework, deep CCA \cite{andrew2013deep} is more powerful to capture nonlinear relationships. Deep canonically correlated autoencoder (DCCAE) \cite{Wang2016On} combines deep CCA and autoencoder structure to learn compact representation. Different from CCA, some approaches \cite{zhao2017multi,zhang2018binary} employ matrix factorization to obtain hierarchical representation from multi-view data with specific constraints. Multi-view dimensionality co-reduction (MDcR) \cite{zhang2017flexible} applies the kernel matching to regularize the dependence across views. Self-representation is also introduced to better incorporate multi-view information \cite{li2019reciprocal,cao2015diversity}. Moreover, generative adversarial network is applied to handle missing view problem \cite{wang2018partial} or impose prior information \cite{tao2019adversarial}. There is a major difference between above approaches and our work - all of them treat each view equally or assign a fixed weight. In contrast, our method considers sample-specific view quality, while the corresponding uncertainty guides a robust multi-view integration.
	
	\paragraph{Data Uncertainty Learning.} Quantifying uncertainty and making reasonable decisions are critical in real-world applications \cite{pate1996uncertainties,faber2005treatment,der2009aleatory}. There are mainly two categories of uncertainty, \textit{data uncertainty} (a.k.a., aleatoric uncertainty) and \textit{model uncertainty} (a.k.a., epistemic uncertainty). Data uncertainty can capture the noise inherent in the observations while model uncertainty (typically in supervised learning) can reflect the prediction confidence \cite{kendall2017uncertainties}. Recently, many research works investigated how to estimate uncertainty in deep learning \cite{blundell2015weight,gal2016dropout}. With these techniques, many computer vision models obtain great improvement on robustness and interpretability. For example, uncertainty modeling is introduced in face recognition \cite{shi2019probabilistic,chang2020data}, and object detection \cite{choi2019gaussian,kraus2019uncertainty}. Some methods \cite{kendall2017uncertainties,kendall2018multi} utilize probability model to capture data uncertainty and reduce the effect of noisy samples. Our method introduces data uncertainty into multi-view learning. With the help of uncertainty, the proposed model can automatically estimate the importances of different views for different samples. Superior performance indicates that incorporating data uncertainty in information integration is more suitable to real-world applications.
	
	\section{Proposed Model}
	Multi-view representation learning (MRL) focuses on learning a unified representation encoding intrinsic information of multiple views. Formally, given a multi-view dataset $\mathcal{X}=\{\mathbf{x}_i^{(1)};...;\mathbf{x}_i^{(V)}\}_{i=1}^N$ which has $V$ different views of observation, the goal of multi-view representation learning is inferring a latent representation $\mathbf{h}$ for each sample. Unfortunately, quality of views usually varies for different samples. For example, in multi-sensor system, there may be corrupted sensors providing inaccurate measurement (high-uncertainty-view), and furthermore there may be samples obtained in unpromising conditions (high-uncertainty-sample). A reliable multi-view representation learning model should take these conditions into consideration. In this section, we will show how to learn reliable representations from multi-view data by capturing the data uncertainty.
	
	\subsection{Uncertainty-Aware Multi-View Integration}
	For real-world applications, data usually contains inevitable noise, which is one of the main challenge in representation learning. In order to model the underlying noise, we assume different observations are sampled from different Gaussian distributions, i.e., $\mathbf{x}_i^{(v)} \sim \mathcal{N}(\boldsymbol{\mu}_i^{(v)}, (\sigma_i^{(v)})^2)$. Accordingly, the observations are modeled as
	\begin{equation}
	\mathbf{x}_i^{(v)}=\boldsymbol{\mu}_i^{(v)}+\epsilon\sigma_i^{(v)}, \epsilon \sim \mathcal{N}(\mathbf{0},\mathbf{I}),
	\end{equation} 
	where the mean variable $\boldsymbol{\mu}_i^{(v)}$ refers to sample identity, and $\sigma_i^{(v)}$ reflects the uncertainty of the observation in the $v^{th}$ view.
	
	Based on above assumption, we target on encoding intrinsic information from multiple views into a unified representation. Considering the unified representation as latent variables, from the perspective of Bayesian, the joint distribution of latent variable $\mathbf{h}_i$ and multiple observations $\mathbf{x}_i^{(v)}$ for $v=1,...,V$ can be decomposed as prior on $\mathbf{h}_i$ ($p(\mathbf{h}_i)$) and likelihood as
	\begin{equation}
	p(\mathbf{x}_i^{(1)},\cdots,\mathbf{x}_i^{(V)},\mathbf{h}_i)=p(\mathbf{x}_i^{(1)},\cdots,\mathbf{x}_i^{(V)}|\mathbf{h}_i)p(\mathbf{h}_i).
	\end{equation}
	Since there is usually no prior knowledge on latent representation, we simply ignore the prior but focus on the likelihood. The likelihood aims to reconstruct observation of each view from the unified representation $\mathbf{h}_i$. The underlying assumption is that observation of each view $\mathbf{x}_i^{(v)}$ is conditionally independent given the latent variable $\mathbf{h}_i$, for which the likelihood can be factorized as
	\begin{equation}
	p(\mathbf{x}_i^{(1)},\cdots,\mathbf{x}_i^{(V)}|\mathbf{h}_i)=p(\mathbf{x}_i^{(1)}|\mathbf{h}_i)\cdots p(\mathbf{x}_i^{(V)}|\mathbf{h}_i).
	\end{equation}
	This implies that we can use multiple neural networks to decode the latent variable into different views. Taking one neural network $f^{(v)}(\cdot)$ for example, we use latent variable $\mathbf{h}_i$ to reconstruct the Gaussian distribution of observation $\mathbf{x}_i^{(v)}$, i.e.,
	\begin{equation}
	p(\mathbf{x}_i^{(v)}|\mathbf{h}_i)=\mathcal{N}(f^{(v)}(\mathbf{h}_i),{(\sigma^{(v)})}^2).
	\end{equation}
	The parameters of neural networks are omitted for simplicity. To capture the uncertainty inherent in each observation instead of fixing it for each view, we model the variance to be variables which can vary with different samples. Then we have
	\begin{equation}
	p(\mathbf{x}_i^{(v)}|\mathbf{h}_i)=\mathcal{N}(f^{(v)}(\mathbf{h}_i),{(g^{(v)}(\mathbf{h}_i))}^2).
	\end{equation}
	Rather than a deterministic output, now we model each observation containing different level of noise. Specifically, the learned variance is a metric that captures the uncertainty caused by noise.
	
	Taking observation $\mathbf{x}_i^{(v)}$ as reconstruction target, it leads to the following likelihood
	\begin{equation}
	\begin{aligned}
	&p(\mathbf{x}_i^{(v)}|\mathbf{h}_i)=\frac{1}{\sqrt{2\pi{(\sigma_i^{(v)})}^2}}\exp(-\frac{||\mathbf{x}_i^{(v)}-\boldsymbol{\mu}_i^{(v)}||^2}{{2(\sigma_i^{(v)})}^2})\\
	&s.t.\quad \boldsymbol{\mu}_i^{(v)}=f^{(v)}(\mathbf{h}_i),~\sigma_i^{(v)}=g^{(v)}(\mathbf{h}_i).
	\end{aligned}
	\end{equation}
	In practice, the log likelihood as follows is maximized
	\begin{equation}
	\ln p(\mathbf{x}_i^{(v)}|\mathbf{h}_i)=-\frac{||\mathbf{x}_i^{(v)}-\boldsymbol{\mu}_i^{(v)}||^2}{{2(\sigma_i^{(v)})}^2}-\ln {(\sigma_i^{(v)})}.
	\end{equation}
	We omit the constant term because it will not affect the optimization. Then, we aim to search a positive scalar $\sigma_i^{(v)}$ for the $v^{th}$ view of the $i^{th}$ sample to weigh the reconstruction loss. The magnitude of variance determines the sharpness of Gaussian distribution. The larger the variance, the higher the uncertainty for the observation. Basically, large uncertainty can always reduce the reconstruction loss, but the second term introduced in the objective acts as a regularizer which constrains the uncertainty from increasing too much and avoids a trivial solution.
	
	The reconstruction networks are utilized to enforce $\mathbf{h}_i$ to contain intrinsic information of multiple views (Fig.~\ref{fig:framework}(a)), which makes it easier to infer $\mathbf{x}_i^{(v)}$. Through this reconstruction process, the latent variable $\mathbf{h}_i$ is optimized along with the parameters of networks. In this way, $\mathbf{h}_i$ is able to reconstruct each view, and thus information from different views can be well encoded into $\mathbf{h}_i$. Note that the flow of information in reconstruction networks is reverse to conventional neural network learning, where the input is observation and the output is latent representation. We term the decoder-like framework (i.e., reconstruction network) as Reversal Network (R-Net).
	
	Accordingly, the final minimization objective of our multi-view model is
	\begin{equation}
	\begin{aligned}
	&\mathcal{L}=\sum_{i=1}^{N}\sum_{v=1}^{V}\left(\frac{||\mathbf{x}_i^{(v)}-\boldsymbol{\mu}_i^{(v)}||^2}{{2(\sigma_i^{(v)})}^2}+\ln {(\sigma_i^{(v)})}\right)\\
	&s.t.\quad \boldsymbol{\mu}_i^{(v)}=f^{(v)}(\mathbf{h}_i),~  \sigma_i^{(v)}=g^{(v)}(\mathbf{h}_i).
	\end{aligned}
	\end{equation}
	The overall proposed model is termed as Dynamic Uncertainty-Aware Networks (DUA-Nets). On the one hand, DUA-Nets estimate uncertainty in multi-view data. Instead of a fixed weight for each view, the model learns input-dependent uncertainty for different samples according to their quality. On the other hand, in DUA-Nets, the latent variable $\mathbf{h}_i$ acts as input and aims to reconstruct the original views in a reversal manner. The uncertainty of each view indicates the possible noise inherent in the observation, and thus it can guide the reconstruction process. With the help of uncertainty, the high-quality samples (and views) can be fully exploited while the effect from the noisy samples (and views) will be alleviated. In this way, we model the noise in multi-view data and reduce its impact to obtain robust representations. The learning process is shown in Fig.~\ref{fig:framework}(b).
	
	\subsection{Why Can Our Model Capture Uncertainty without Supervision?} 
	There may be a natural question: since most existing models estimate uncertainty with the help of class labels, how can we learn the uncertainty inherent in data without supervision? First, if given noise-free data, our model is able to promisingly reconstruct each observation. In this case, the estimated uncertainty is close to zero. While with noise in the data, the reconstruction loss will increase accordingly. The principle behind this is that the real-world data distributions have natural patterns that neural networks can easily capture. The noisy signals are usually high-frequency components that are difficult to model. Thus, it is difficult for neural networks to reconstruct noisy data, which causes large reconstruction loss on these samples. The assumption is consistent with prior study \cite{ulyanov2018deep}. Therefore, when the low-quality data is as input, our model tends to output a larger reconstruction loss, and the corresponding uncertainty will be larger to prevent the reconstruction loss from increasing too much. Each R-Net is able to capture the data noise in each view, and further assigns different views with corresponding weights to produce a unified representation. We will show this effect in the experiments section (Fig. \ref{fig:Noise_rate}).
	
	
	\section{Experiments}
	In the experiments, we evaluate the proposed algorithm on real-world multi-view datasets and compare it with existing multi-view representation learning methods. The learned representation is evaluated by conducting clustering and classification tasks. Furthermore, we also provide the analysis of uncertainty estimation and robustness evaluation on noisy data.
	
	\begin{table*}[!ht]
		\small
		\centering
		\begin{tabular}{clccccccc}
			\toprule
			dataset & metric & DCCA  & DCCAE  & MDcR & DMF-MVC & RMSL  & R-Nets & DUA-Nets \\
			\midrule[1pt]
			\multirow{4}[0]{*}{UCI-MF}
			& ACC  & 66.26 $\pm$ 0.16  & 69.17 $\pm$ 1.02   & 76.72 $\pm$ 2.77   & 71.86 $\pm$ 4.25  & 77.07 $\pm$ 5.36   & 94.64 $\pm$ 4.38   & \textbf{96.50 $\pm$ 0.81}  \\
			& NMI  & 66.01 $\pm$ 0.45  & 66.96 $\pm$ 0.91   & 76.68 $\pm$ 0.93   & 73.09 $\pm$ 3.23  & 75.54 $\pm$ 3.09   & 91.70 $\pm$ 2.47  & \textbf{93.04 $\pm$ 0.65}  \\
			& F    & 59.05 $\pm$ 0.39  & 60.50 $\pm$ 1.10   & 71.93 $\pm$ 2.22   & 66.66 $\pm$ 4.69  & 66.43 $\pm$ 6.77   & 92.01 $\pm$ 3.55  & \textbf{93.75 $\pm$ 0.75}  \\
			& RI   & 91.39 $\pm$ 0.06  & 91.77 $\pm$ 0.21   & 94.11 $\pm$ 0.48   & 92.85 $\pm$ 1.13  & 92.08 $\pm$ 2.57   & 98.30 $\pm$ 0.89  & \textbf{98.72 $\pm$ 0.17} \\
			\hline
			\multirow{4}[0]{*}{ORL}
			& ACC  & 59.68 $\pm$ 2.04  & 59.40 $\pm$ 2.20   & 61.70 $\pm$ 2.19   & 65.38 $\pm$ 2.86  & \textbf{76.95 $\pm$ 1.95}   & 68.85 $\pm$ 2.16   & 70.38 $\pm$ 1.25  \\
			& NMI  & 77.84 $\pm$ 0.83  & 77.52 $\pm$ 0.86   & 79.45 $\pm$ 1.20   & 82.87 $\pm$ 1.26  & \textbf{91.29 $\pm$ 1.30}   & 84.05 $\pm$ 0.77   & 85.49 $\pm$ 0.76  \\
			& F    & 47.72 $\pm$ 2.05  & 46.71 $\pm$ 2.22   & 48.48 $\pm$ 2.59   & 52.01 $\pm$ 3.43  & 65.30 $\pm$ 2.17   & 66.72 $\pm$ 1.51   & \textbf{69.32 $\pm$ 1.39}  \\
			& RI   & 97.42 $\pm$ 0.13  & 97.39 $\pm$ 0.14   & 97.28 $\pm$ 0.22   & 97.29 $\pm$ 0.30  & 97.96 $\pm$ 0.63   & 97.90 $\pm$ 0.14   & \textbf{98.06 $\pm$ 0.09}  \\
			\hline
			\multirow{4}[0]{*}{COIL20}
			& ACC  & 63.73 $\pm$ 0.78  & 62.72 $\pm$ 1.40   & 64.25 $\pm$ 2.98   & 53.92 $\pm$ 5.89  & 63.04 $\pm$ 2.20   & 66.47 $\pm$ 6.73   & \textbf{72.28 $\pm$ 4.79}  \\
			& NMI  & 76.02 $\pm$ 0.50  & 76.32 $\pm$ 0.66   & 79.44 $\pm$ 1.37   & 72.36 $\pm$ 2.11  & 79.24 $\pm$ 1.91   & 79.56 $\pm$ 2.23   & \textbf{82.72 $\pm$ 1.81}  \\
			& F    & 58.76 $\pm$ 0.53  & 57.56 $\pm$ 1.15   & 63.60 $\pm$ 2.57   & 46.39 $\pm$ 4.97  & 61.05 $\pm$ 2.79   & 66.48 $\pm$ 4.22   & \textbf{71.47 $\pm$ 3.35}  \\
			& RI   & 95.60 $\pm$ 0.06  & 95.27 $\pm$ 0.30   & 96.11 $\pm$ 0.29   & 92.56 $\pm$ 1.46  & 95.67 $\pm$ 0.61   & 96.03 $\pm$ 0.55   & \textbf{96.77 $\pm$ 0.49}  \\
			\hline
			\multirow{4}[0]{*}{MSRCV1}
			& ACC  & 76.09 $\pm$ 0.08  & 65.43 $\pm$ 3.94   & 80.81 $\pm$ 2.13 & 32.19 $\pm$ 1.64  & 69.71 $\pm$ 1.25   & 81.52 $\pm$ 3.65   & \textbf{84.67 $\pm$ 3.03} \\
			& NMI  & 66.04 $\pm$ 0.09  & 60.75 $\pm$ 2.83   & 72.58 $\pm$ 1.94 & 17.32 $\pm$ 2.22  & 59.78 $\pm$ 0.91   & 76.77 $\pm$ 3.19   & \textbf{77.26 $\pm$ 2.80} \\
			& F    & 62.38 $\pm$ 0.12  & 55.27 $\pm$ 3.46   & \textbf{80.83 $\pm$ 2.34} & 23.45 $\pm$ 2.73  & 53.61 $\pm$ 1.22   & 75.10 $\pm$ 3.30   & 76.85 $\pm$ 3.17 \\
			& RI   & 89.62 $\pm$ 0.07  & 85.33 $\pm$ 1.63   & 91.05 $\pm$ 0.60 & 70.39 $\pm$ 2.19  & 86.08 $\pm$ 0.89   & 92.52 $\pm$ 0.99   & \textbf{93.18 $\pm$ 0.99} \\
			\hline
			\multirow{4}[0]{*}{CUB}  
			& ACC  & 54.50 $\pm$ 0.29  & 66.70 $\pm$ 1.52   & \textbf{73.68 $\pm$ 3.32}   & 37.50 $\pm$ 2.45  & 66.47 $\pm$ 4.58   & 70.53 $\pm$ 2.03   & 72.98 $\pm$ 2.97  \\
			& NMI  & 52.53 $\pm$ 0.19  & 65.76 $\pm$ 1.36   & \textbf{74.49 $\pm$ 0.75}   & 37.82 $\pm$ 2.04  & 68.95 $\pm$ 4.28   & 71.63 $\pm$ 1.65   & 72.89 $\pm$ 1.59  \\
			& F    & 45.84 $\pm$ 0.31  & 58.22 $\pm$ 1.18   & 65.72 $\pm$ 1.37   & 28.95 $\pm$ 1.54  & 57.58 $\pm$ 6.93   & 64.79 $\pm$ 2.25   & \textbf{66.33 $\pm$ 2.01}  \\
			& RI   & 88.61 $\pm$ 0.06  & 91.27 $\pm$ 0.24   & 92.75 $\pm$ 0.44   & 85.52 $\pm$ 0.26  & 89.76 $\pm$ 2.88 	& 92.25 $\pm$ 0.50   & \textbf{92.79 $\pm$ 0.42}  \\
			\hline
			\multirow{4}[0]{*}{Caltech101}
			& ACC  &  49.85 $\pm$ 6.93 & 53.93 $\pm$ 5.78   & 46.51 $\pm$ 0.67   & 52.75 $\pm$ 5.67  & 53.13 $\pm$ 9.63    & 50.86 $\pm$ 2.75   & \textbf{54.55 $\pm$ 0.68}  \\
			& NMI  & 49.51 $\pm$ 5.18  & 53.94 $\pm$ 3.73   & \textbf{56.43 $\pm$ 0.56}   & 45.52 $\pm$ 2.28  & 23.96 $\pm$ 9.03    & 52.31 $\pm$ 0.87   & 49.38 $\pm$ 0.91  \\
			& F    & 56.31 $\pm$ 8.44  & 57.57 $\pm$ 7.08   & 51.55 $\pm$ 0.56   & 55.67 $\pm$ 5.50  & 48.03 $\pm$ 3.04    & 57.62 $\pm$ 2.36   & \textbf{58.37 $\pm$ 0.36}  \\
			& RI   & 72.79 $\pm$ 3.78  & 74.12 $\pm$ 3.27   & 73.27 $\pm$ 0.30   & 73.43 $\pm$ 2.33  & 57.77 $\pm$ 8.02    & 74.57 $\pm$ 0.63   & \textbf{74.68 $\pm$ 0.16}  \\
			\bottomrule
		\end{tabular}%
		\caption{Results of representation clustering performance.}
		\label{tab:clustering}
	\end{table*}%
	
	\subsection{Datasets}
	We conduct experiments on six real-world multi-view datasets as follows: \textbf{UCI-MF} (UCI Multiple Features)\footnote{https://archive.ics.uci.edu/ml/datasets/Multiple+Features}: This dataset consists of handwritten numerals (`0'--`9') from a collection of Dutch utility maps. These digits are represented with six types of features.  \textbf{ORL}\footnote{http://www.cl.cam.ac.uk/research/dtg/attarchive/facedatabase.html}: ORL face dataset contains 10 different images of each of 40 distinct subjects under different conditions. Three types of features: intensity, LBP and  Gabor are used as different views.
	\textbf{COIL20MV}\footnote{http://www.cs.columbia.edu/CAVE/software/softlib/}: There are 1440 images from 20 object categories. Three types of features that are same to ORL are used.
	\textbf{MSRCV1} \cite{xu2016discriminatively}: This dataset contains 30 different images for each class out of 7 classes in total. Six types of features: CENT, CMT, GIST, HOG, LBP, SIFT are extracted.
	\textbf{CUB}\footnote{http://www.vision.caltech.edu/visipedia/CUB-200.html}: Caltech-UCSD Birds dataset contains 200 different bird	categories with 11788 images and text descriptions. Features of 10 categories are extracted by GoogLeNet and Doc2Vec in Gensim\footnote{https://radimrehurek.com/gensim/models/doc2vec.html}.
	\textbf{Caltech101}\footnote{http://www.vision.caltech.edu/Image\_Datasets/Caltech101}: This dataset contains images of 101 object categories. About 40 to 800 images per category. We use a subset of 1,474 images with 6 views.
	
	\subsection{Compared Methods}
	We compare the proposed DUA-Nets with multi-view representation learning methods as follows: 
	\begin{itemize}
		\item \textbf{DCCA}: Deep Canonically Correlated Analysis \cite{andrew2013deep} extends CCA \cite{hotelling1936relations} by applying deep neural networks to learn nonlinear projection. DCCA maximizes the correlation between learned representations of two views.
		\item \textbf{DCCAE}: Deep Canonically Correlated AutoEncoders \cite{Wang2016On} employs autoencoders structure to obtain better embedding.
		\item \textbf{MDcR}: Multi-view Dimensionality co-Reduction
		\cite{zhang2017flexible} applies kernel matching constraint to enhance correlations among multiple views and combine these projected low-dimensional features together.
		\item \textbf{DMF-MVC}: Deep Semi-Non-negative Matrix Factorization for Multi-View Clustering \cite{zhao2017multi} uses deep neural networks to conduct semi-NMF on multi-view data and seek the consistent representation.
		\item \textbf{RMSL}: Reciprocal Multi-layer Subspace Learning \cite{li2019reciprocal} uses self representation and reciprocal encoding to explore the consistency and complementary information among multiple views.
	\end{itemize}
	
	\subsection{Implementation Details}
	There are two parts in the proposed DUA-Nets, view-specific reconstruction network and uncertainty estimation network. We employ similar network architecture for both components. For all datasets, 3-layer fully connected network followed by ReLU activation function is used in the experiments. The latent representation $\mathbf{h}_i$ is randomly initialized with Gaussian distribution. Adam optimizer \cite{kingma2014adam} is employed for optimization of all parameters. The model is implemented by PyTorch on one NVIDIA Geforce GTX TITAN Xp with GPU of 12GB memory.
	\begin{table*}[!ht]
		\small
		\centering
		\setlength{\tabcolsep}{1.7mm}{
		\begin{tabular}{ccccccccc}
			\toprule
			dataset & proportion & DCCA  & DCCAE  & MDcR & DMF-MVC & RMSL  & R-Nets & DUA-Nets \\
			\midrule[1pt]
			\multirow{4}[0]{*}{UCI-MF} 
			& $G_{8}/P_{2}$       & 95.18 $\pm$ 0.55    & 95.78 $\pm$ 0.46   & 92.33 $\pm$ 0.73   & 94.68 $\pm$ 0.71    & 93.05 $\pm$ 1.17   & 96.78 $\pm$ 0.38   & \textbf{98.10 $\pm$ 0.32}  \\
			& $G_{7}/P_{3}$       & 94.62 $\pm$ 0.64    & 95.10 $\pm$ 0.64   & 91.55 $\pm$ 0.39   & 93.72 $\pm$ 0.60    & 91.67 $\pm$ 1.14   & 96.55 $\pm$ 0.41   & \textbf{97.98 $\pm$ 0.47}  \\
			& $G_{5}/P_{5}$       & 94.35 $\pm$ 0.46    & 94.79 $\pm$ 0.58   & 91.41 $\pm$ 0.68   & 93.33 $\pm$ 0.46    & 90.74 $\pm$ 1.32   & 95.95 $\pm$ 0.59   & \textbf{97.72 $\pm$ 0.15}  \\
			& $G_{2}/P_{8}$       & 92.79 $\pm$ 0.51    & 92.63 $\pm$ 0.54   & 88.11 $\pm$ 0.61   & 88.23 $\pm$ 0.57    & 88.60 $\pm$ 0.78   & 93.34 $\pm$ 0.62   & \textbf{96.44 $\pm$ 0.46}  \\
			\hline
			\multirow{4}[0]{*}{ORL}
			& $G_{8}/P_{2}$       & 83.25 $\pm$ 2.71    & 81.62 $\pm$ 2.95   & 92.00 $\pm$ 1.58   & 93.13 $\pm$ 1.21    & \textbf{96.37 $\pm$ 1.38}   & 93.88 $\pm$ 2.05   & 94.14 $\pm$ 0.57  \\
			& $G_{7}/P_{3}$       & 78.92 $\pm$ 1.93    & 80.00 $\pm$ 1.47   & 90.83 $\pm$ 2.08   & 91.75 $\pm$ 1.64    & \textbf{95.83 $\pm$ 1.47}   & 91.75 $\pm$ 1.51   & 92.94 $\pm$ 1.03  \\
			& $G_{5}/P_{5}$       & 71.15 $\pm$ 1.86    & 72.80 $\pm$ 2.04   & 83.35 $\pm$ 1.08   & 85.45 $\pm$ 1.85    & \textbf{94.30 $\pm$ 1.64}   & 85.22 $\pm$ 1.62   & 86.36 $\pm$ 0.79  \\
			& $G_{2}/P_{8}$       & 51.69 $\pm$ 1.75    & 51.25 $\pm$ 1.90   & 57.38 $\pm$ 2.08   & 56.44 $\pm$ 2.50    & \textbf{84.63 $\pm$ 1.14}   & 57.23 $\pm$ 1.79   & 59.56 $\pm$ 1.05  \\
			\hline
			\multirow{4}[0]{*}{COIL20} 
			& $G_{8}/P_{2}$       & 90.96 $\pm$ 1.24    & 92.54 $\pm$ 0.70   & 91.11 $\pm$ 0.80   & 95.25 $\pm$ 1.06    & 93.14 $\pm$ 1.60   & 99.44 $\pm$ 0.09   & \textbf{99.65 $\pm$ 0.27}  \\
			& $G_{7}/P_{3}$       & 90.48 $\pm$ 1.56    & 91.88 $\pm$ 1.44   & 90.29 $\pm$ 1.05   & 94.76 $\pm$ 0.77    & 91.79 $\pm$ 1.43   & 99.28 $\pm$ 0.41   & \textbf{99.42 $\pm$ 0.43}  \\
			& $G_{5}/P_{5}$       & 88.65 $\pm$ 0.84    & 90.35 $\pm$ 0.58   & 87.63 $\pm$ 1.12   & 92.07 $\pm$ 0.61    & 90.32 $\pm$ 1.24   & 97.31 $\pm$ 0.52   & \textbf{98.67 $\pm$ 0.23}  \\
			& $G_{2}/P_{8}$       & 83.35 $\pm$ 0.60    & 84.11 $\pm$ 1.10   & 79.46 $\pm$ 1.39   & 82.96 $\pm$ 1.03    & 85.65 $\pm$ 1.01   & 87.49 $\pm$ 0.99   & \textbf{92.51 $\pm$ 0.44}  \\
			\hline
			\multirow{4}[0]{*}{MSRCV1} 
			& $G_{8}/P_{2}$       & 79.52 $\pm$ 4.52    & 72.62 $\pm$ 4.38   & \textbf{85.25 $\pm$ 2.21}   & 41.67 $\pm$ 4.52    & 79.52 $\pm$ 3.58   & 79.00 $\pm$ 2.24 & 82.40 $\pm$ 2.08 \\
			& $G_{7}/P_{3}$       & 76.90 $\pm$ 2.26    & 70.55 $\pm$ 5.67	 & \textbf{84.10 $\pm$ 3.17}   & 36.67 $\pm$ 4.16    & 78.25 $\pm$ 2.90   &78.10 $\pm$ 3.16 & 81.75 $\pm$ 3.42	\\
			& $G_{5}/P_{5}$       & 65.05 $\pm$ 0.90	& 68.89 $\pm$ 2.77	 & 79.86 $\pm$ 2.97   & 35.05 $\pm$ 2.27    & 77.90 $\pm$ 1.89   & 77.33 $\pm$ 2.65 & \textbf{80.86 $\pm$ 2.43} \\
			& $G_{2}/P_{8}$       & 43.69 $\pm$ 1.13	& 59.85 $\pm$ 3.40 	 & 72.55 $\pm$ 2.54   & 28.81 $\pm$ 1.55    & 71.69 $\pm$ 2.51   & 70.67 $\pm$ 1.14 & \textbf{73.51 $\pm$ 3.33} \\
			\hline
			\multirow{4}[0]{*}{CUB} 
			& $G_{8}/P_{2}$       & 65.67 $\pm$ 2.85    & 77.00 $\pm$ 2.94   & 79.08 $\pm$ 3.43   & 60.08 $\pm$ 2.79    & 78.70 $\pm$ 2.50   & 75.73 $\pm$ 0.91   & \textbf{80.25 $\pm$ 2.98}  \\
			& $G_{7}/P_{3}$       & 64.83 $\pm$ 1.83    & 74.56 $\pm$ 2.74   & 78.44 $\pm$ 3.08   & 58.56 $\pm$ 2.84    & 77.61 $\pm$ 1.38   & 74.11 $\pm$ 1.51   & \textbf{79.67 $\pm$ 0.65}  \\
			& $G_{5}/P_{5}$       & 62.37 $\pm$ 1.58    & 72.60 $\pm$ 2.52   & 77.53 $\pm$ 1.67   & 55.30 $\pm$ 1.90    & 75.48 $\pm$ 1.57   & 72.48
			$\pm$ 0.87   & \textbf{77.87 $\pm$ 2.14}  \\
			& $G_{2}/P_{8}$       & 58.44 $\pm$ 2.92    & 67.35 $\pm$ 3.84   & \textbf{74.58 $\pm$ 1.65}   & 49.60 $\pm$ 1.38    & 70.35 $\pm$ 1.95   & 62.77 $\pm$ 1.88   & 68.17 $\pm$ 1.44  \\
			\hline
			\multirow{4}[0]{*}{Caltech101} 
			& $G_{8}/P_{2}$       & 92.12 $\pm$ 0.58    & 91.58 $\pm$ 1.02   & 90.14 $\pm$ 0.74   & 85.51 $\pm$ 1.05    & 40.71 $\pm$ 3.08   & 92.81 $\pm$ 0.66   & \textbf{93.63 $\pm$ 0.58}  \\
			& $G_{7}/P_{3}$       & 91.46 $\pm$ 0.70    & 90.91 $\pm$ 0.75   & 89.45 $\pm$ 0.76   & 84.67 $\pm$ 0.82    & 39.76 $\pm$ 1.74   & 92.23 $\pm$ 0.42   & \textbf{93.16 $\pm$ 0.45}  \\
			& $G_{5}/P_{5}$       & 91.30 $\pm$ 0.48    & 90.54 $\pm$ 0.44   & 88.95 $\pm$ 0.41   & 81.88 $\pm$ 0.73    & 37.14 $\pm$ 1.22   & 91.42 
			$\pm$ 0.21   & \textbf{92.18 $\pm$ 0.52}  \\
			& $G_{2}/P_{8}$       & 88.73 $\pm$ 0.38    & 89.44 $\pm$ 0.43   & 88.46 $\pm$ 0.35   & 74.19 $\pm$ 0.99    & 33.82 $\pm$ 1.36   & 88.51 $\pm$ 0.48   & \textbf{89.72 $\pm$ 0.77}  \\
			\bottomrule[1.5pt]
		\end{tabular}}
		\caption{Performance comparison on classification task.}    
		\label{tab:classification}%
	\end{table*}%

	\subsection{Performance Evaluation on Clustering}
	We apply DUA-Nets and compared methods to learn multi-view representations, then we conduct clustering task to evaluate these learned representations. We employ k-means algorithm because it is simple and intuitive that can well reflect the structure of representations. For quantitative comparison of these methods, we use four common evaluation metrics including accuracy (ACC), normalized mutual information (NMI), rand index (RI), F-score to comprehensively evaluate different properties of clustering results. For each of these metrics, a higher value indicates a better performance. In order to reduce the impact of randomness, we run each method for 30 times. Clustering results are shown in Table~\ref{tab:clustering}. We report the performance of our method without uncertainty (R-Nets) as an ablation comparison. It is observed that DUA-Nets achieve better performance on most datasets. Take UCI-MF for example, R-Nets improve 17\% over RMSL. DUA-Nets further outperform R-Nets, which validates that our modeling of uncertainty can capture the noise inherent in data and further promote representation learning.
	
	\subsection{Performance Evaluation on Classification}
	We also conduct experiments on classification task based on the learned representations. KNN algorithm is used for its simplicity. We divide the learned representations into training and testing sets with different proportions, denoted as $G_{ratio}/P_{ratio}$, where $G$ is ``gallery set" and $P$ is ``probe set". Accuracy (ACC) is used as evaluation metric. We run 30 times for each partition and report the mean and standard deviation value as well. 	Classification results are shown in Table~\ref{tab:classification}. DUA-Nets achieve more promising performance than compared methods. RMSL and MDcR perform better on some cases but DUA-Nets show more stable performance on all cases.
	
	\begin{figure}[htp]
		\centering 
		\subfigure[$\eta = 0.1$.]{
			\includegraphics[width=4.2cm]{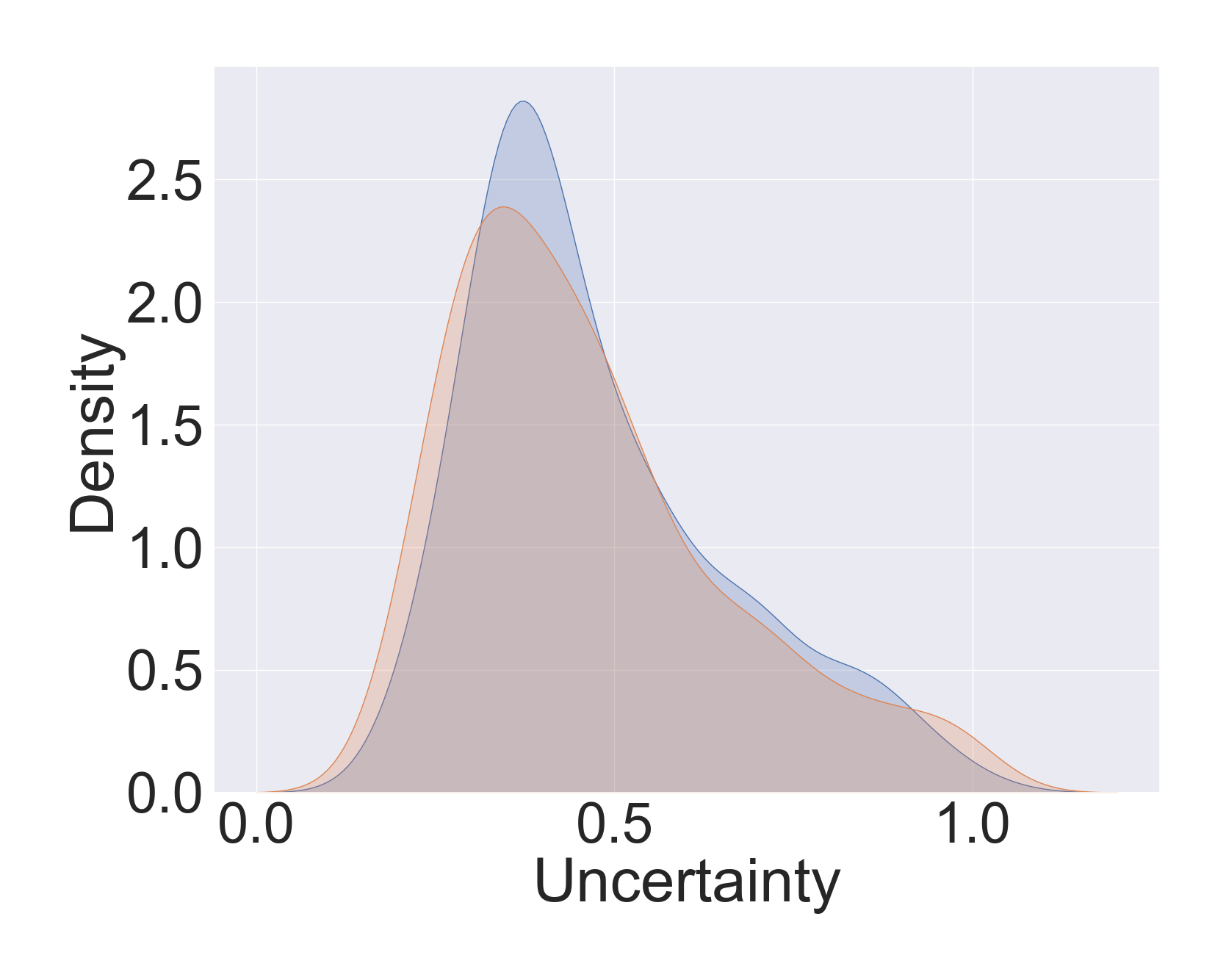}
		}\hspace{-5mm}
		\subfigure[$\eta = 0.5$.]{
			\includegraphics[width=4.2cm]{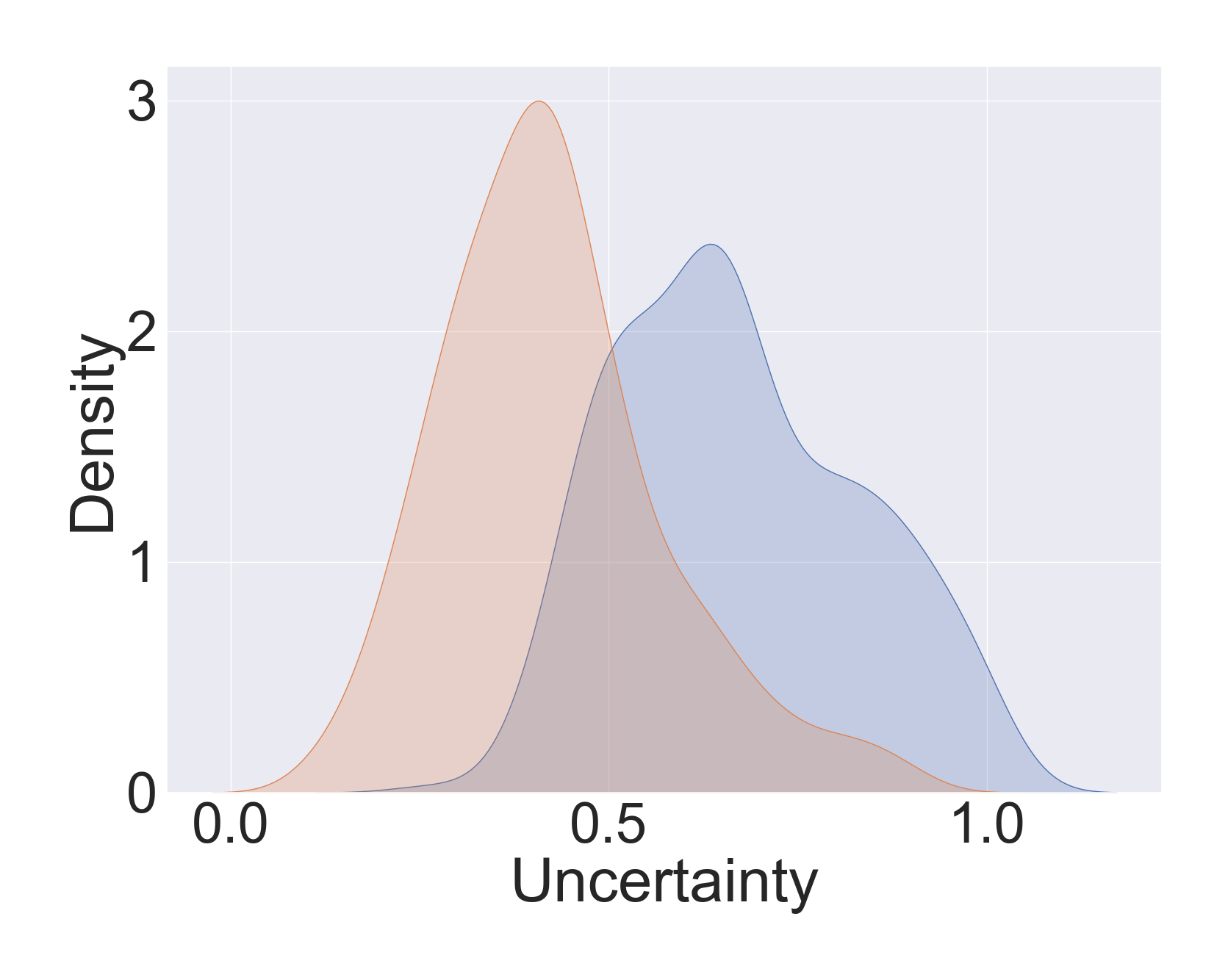}
		}\hspace{-5mm}
		\subfigure[$\eta = 1$.]{
			\includegraphics[width=4.2cm]{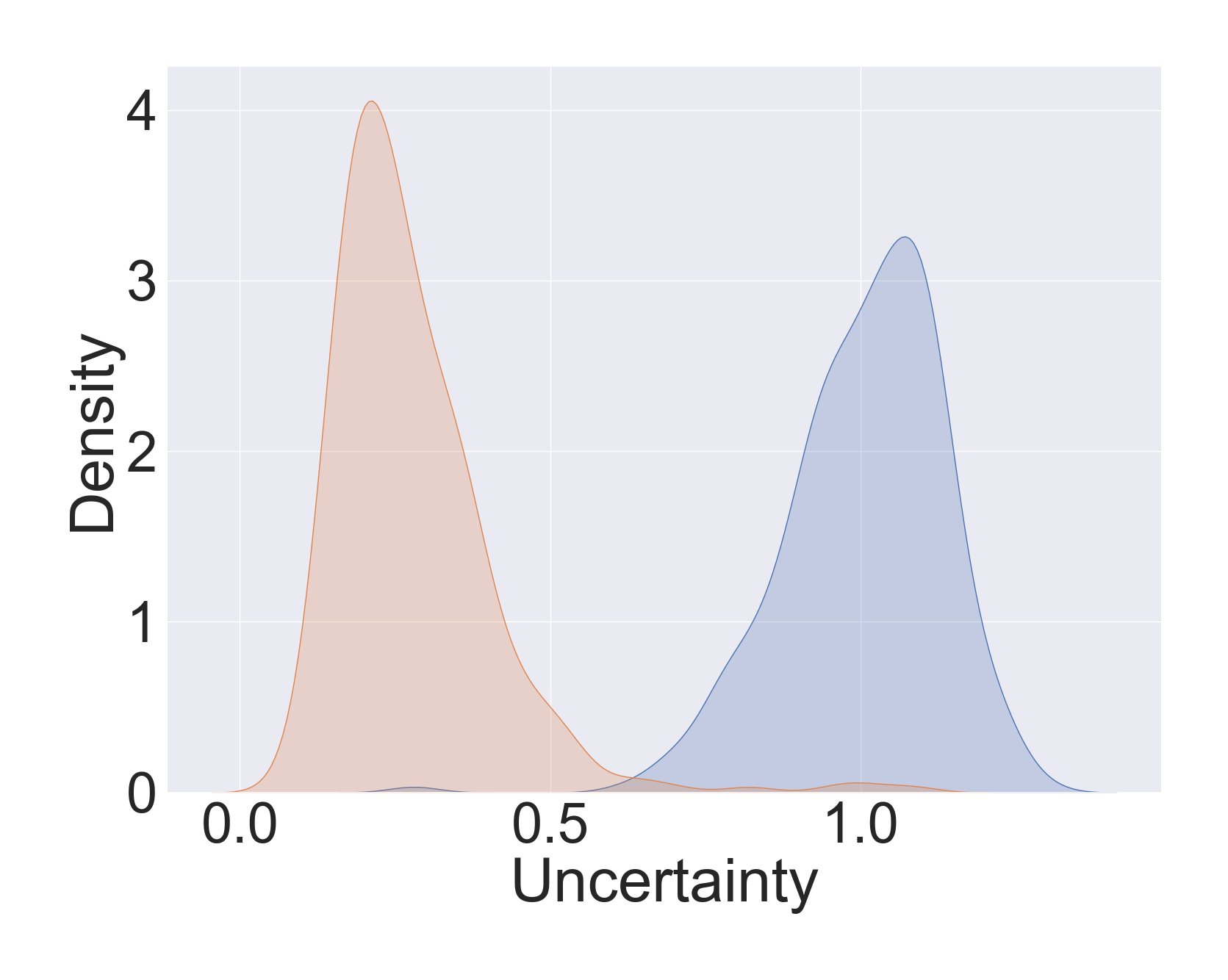}
		}\hspace{-5mm}
		\subfigure[$\eta = 2$.]{
			\includegraphics[width=4.2cm]{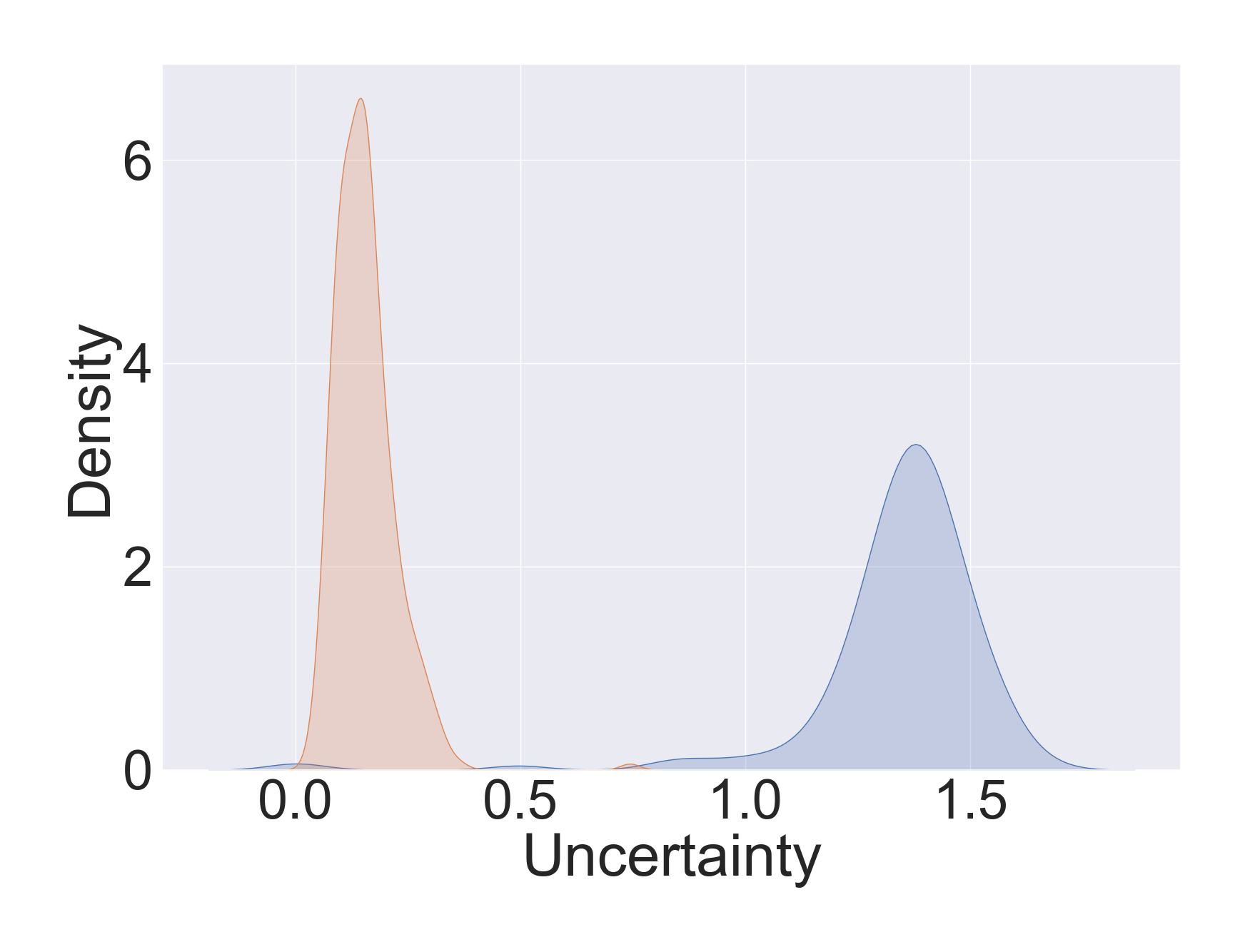}
		}
		\caption{Investigation of our model in capturing data noise. The curves in blue and orange correspond to distributions of noisy and clean data, respectively. The uncertainty basically becomes larger with the increasing of noise intensity.}
		\label{fig:Noise_degree_uncertainty}
	\end{figure}
	
	\subsection{Uncertainty Estimation Analysis}
	According to the comparison between R-Nets and DUA-Nets, it seems the uncertainty is critical for the improvement. Therefore, in this part, we conduct qualitative experiments to provide some insights for the estimated uncertainty.
	
	\textbf{Ability of capturing uncertainty.}
	We estimate data uncertainty on the modified CUB dataset. There are two views in the dataset, we add noise to half of the data in one view. Specifically, we generate $\frac{N}{2}$ noise vectors that are sampled from Gaussian distribution $\mathcal{N}(\mathbf{0}, \mathbf{I})$. Then, we add these noise vectors (denoted as $\boldsymbol{\epsilon}$) multiplied with intensity $\eta$ to pollute half of the original data, i.e., $\mathbf{\tilde{x}}_i^{(1)} = \mathbf{x}_i^{(1)}+\eta \boldsymbol{\epsilon}_i$, for $i=1,...,\frac{N}{2}$. The Gaussian kernel density estimation \cite{scott2015multivariate} of learned uncertainty is shown in Fig.~\ref{fig:Noise_degree_uncertainty}. It can be found that when the noise intensity is small ($\eta=0.1$), the distribution curves of noisy samples and clean samples are highly overlapped. With the increasing of noise intensity, the uncertainty of noisy samples grows correspondingly. This demonstrates that the estimated uncertainty is closely related to sample quality, which validates that the proposed DUA-Nets are aware of the quality of each observation, and guide the integration of views into promising representations.

\begin{figure}[]
		\centering
		\subfigure[View 1: Noise ratio 10\%.]{
			\includegraphics[width=4.0cm]{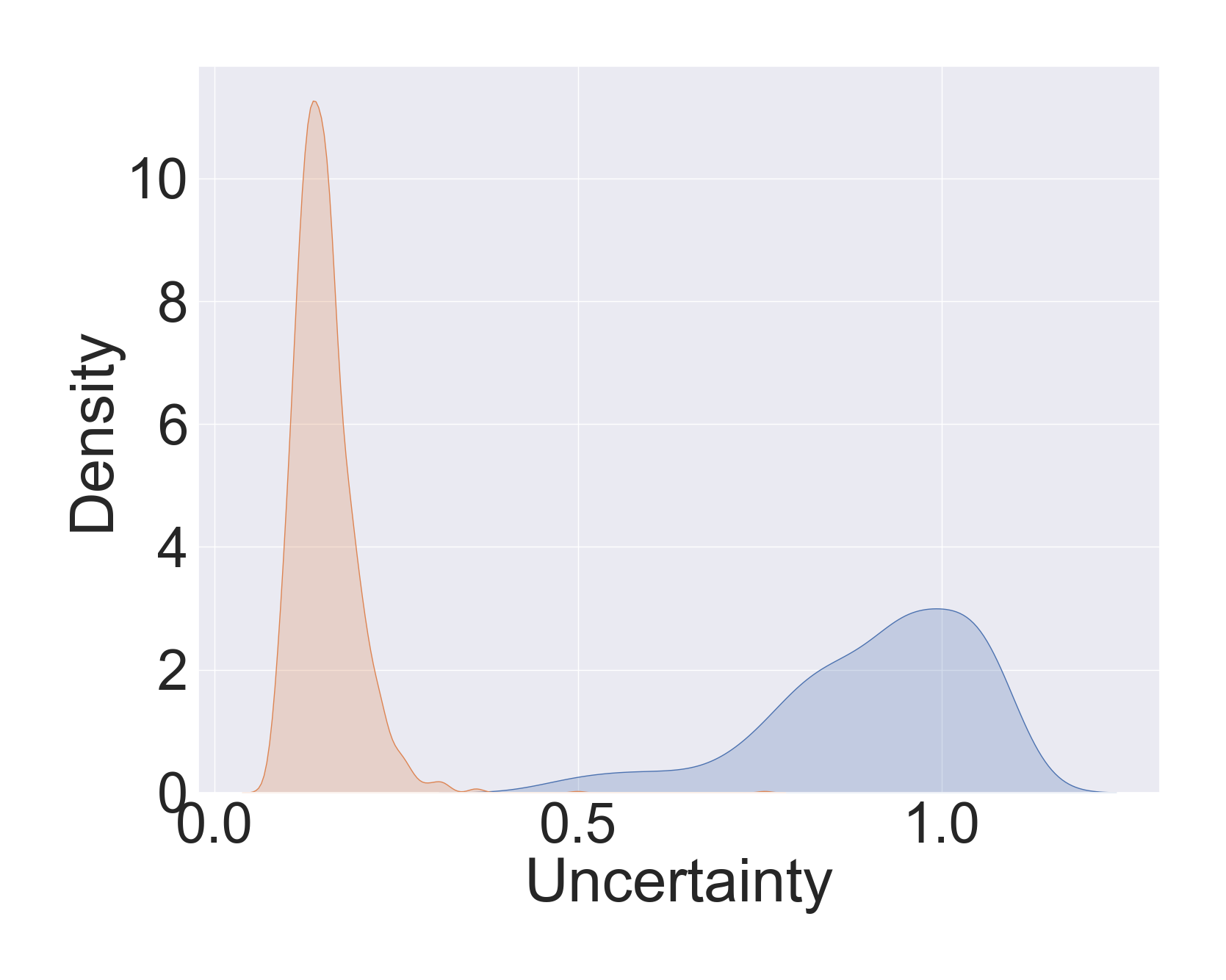}
		}
		\hspace{-5mm}
		\subfigure[View 1: Noise ratio 50\%.]{
			\includegraphics[width=4.0cm]{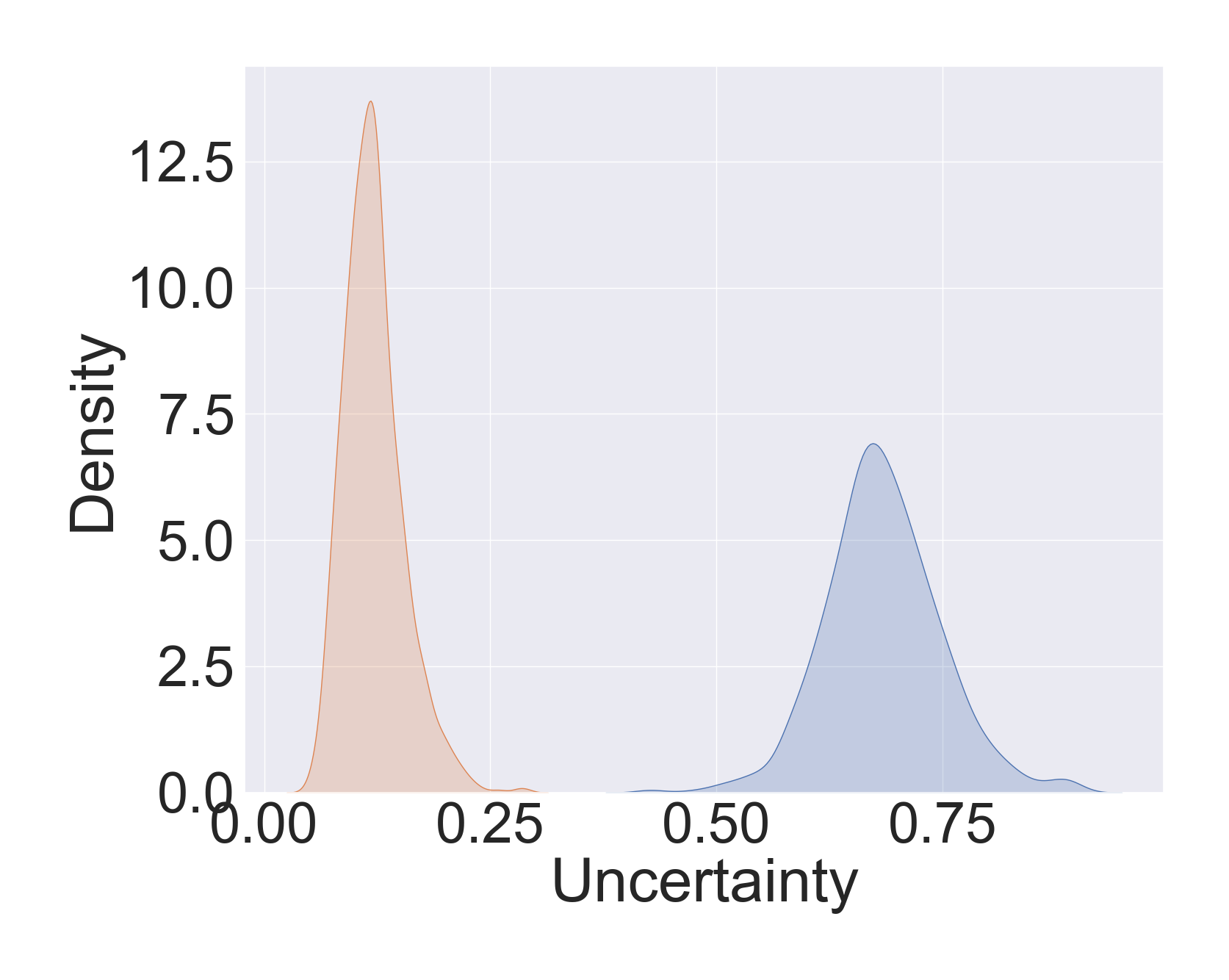}
		}
		\hspace{-5mm}
		\subfigure[View 2: Noise ratio 10\%.]{
			\includegraphics[width=4.0cm]{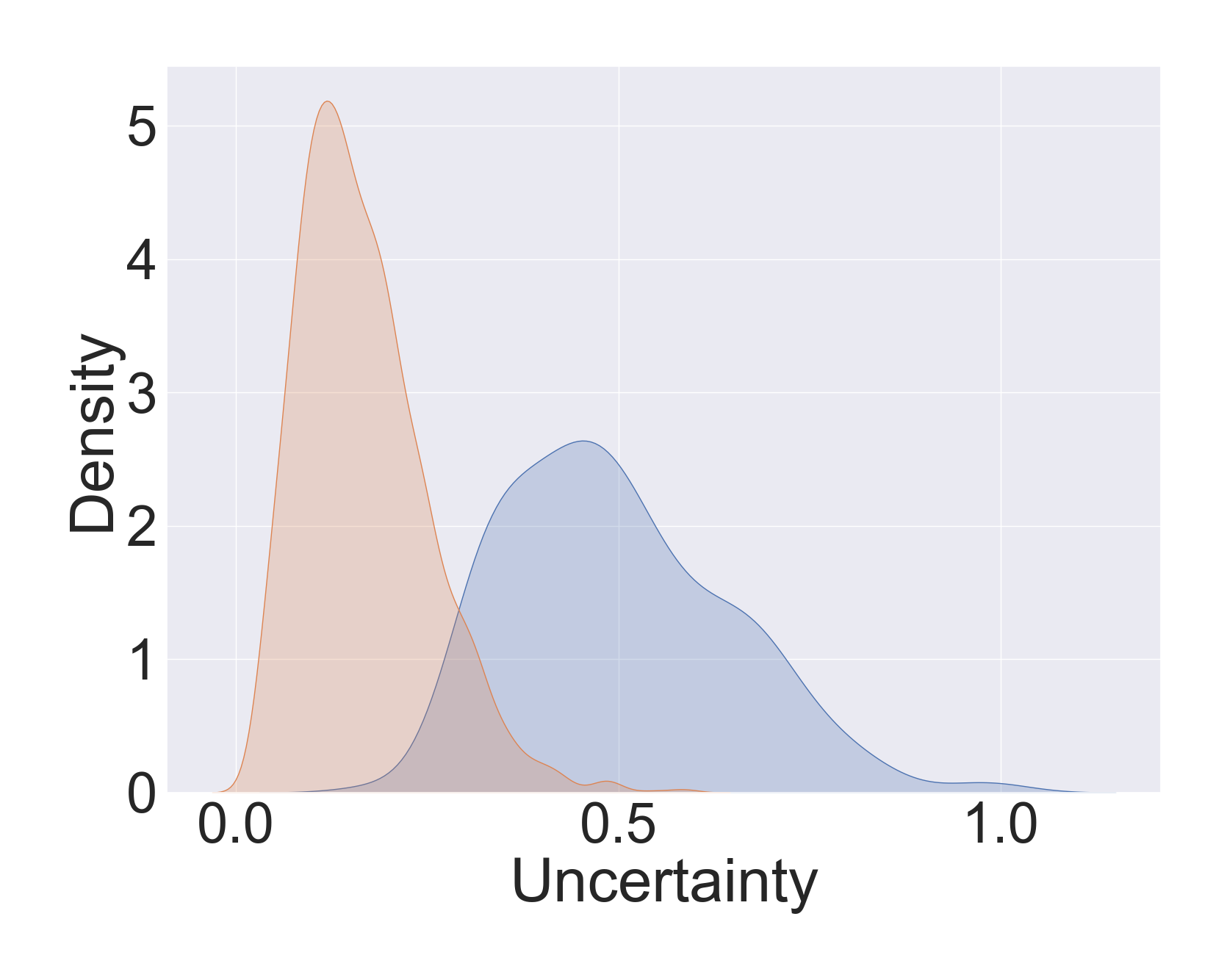}
		}
		\hspace{-5mm}
		\subfigure[View 2: Noise ratio 50\%.]{
			\includegraphics[width=4.0cm]{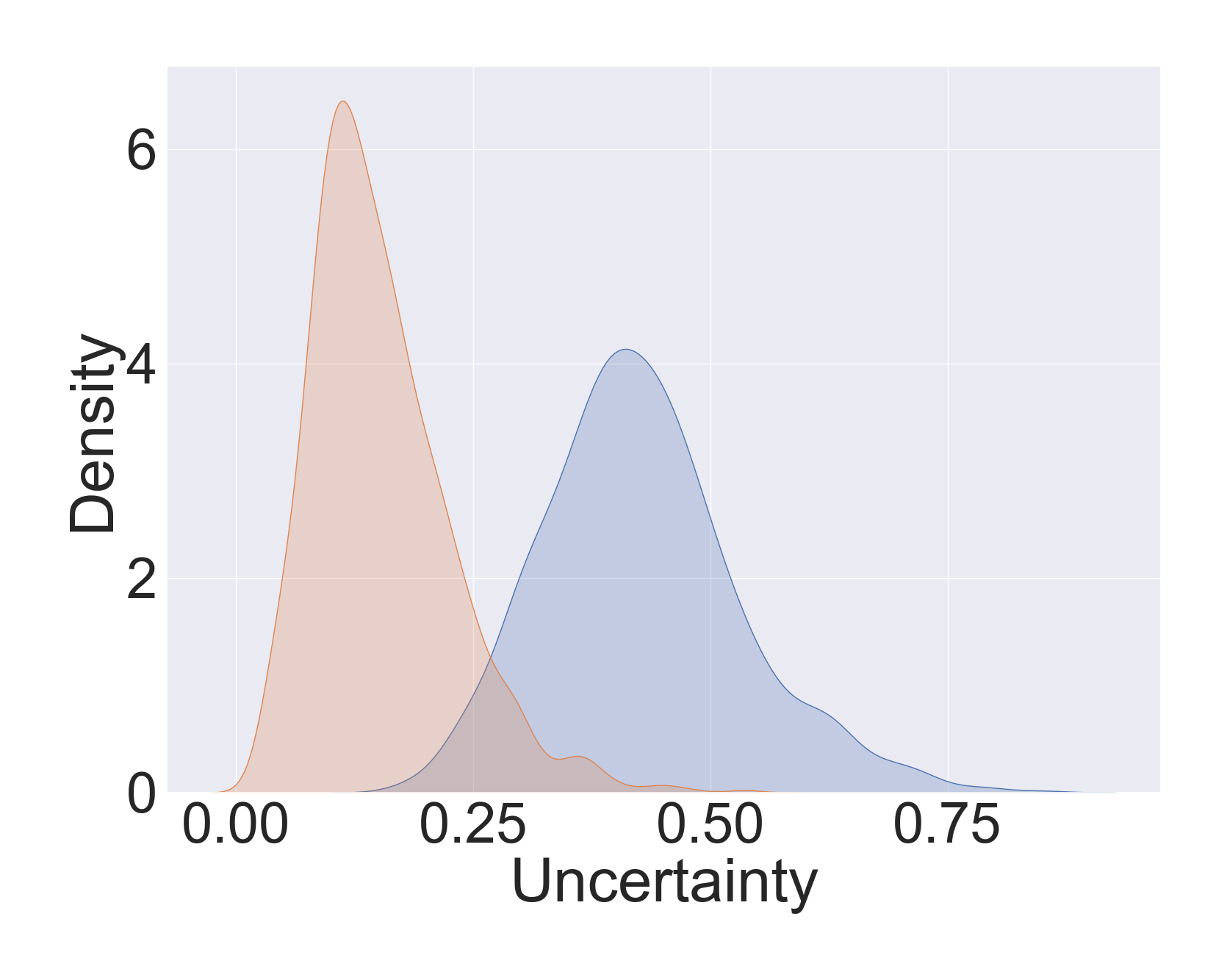}
		}
		\caption{Investigation of ``Why can our model capture uncertainty without supervision?" The curves in blue and orange correspond to distributions of noisy and clean data, respectively. The model is able to capture uncertainty even with large ratio of noisy data.}
		\label{fig:Noise_rate}
	\end{figure}

	\textbf{Capture uncertainty without supervision.}
	In order to investigate the principle behind uncertainty estimation in unsupervised manner, we synthetically add Gaussian noise to different ratio of samples. Typically, we conduct view-specific uncertainty estimation on first two views of UCI-MF dataset individually. Fig.~\ref{fig:Noise_rate} shows the Gaussian kernel density estimation of estimated uncertainty. DUA-Nets are able to capture uncertainty of observations. As the number of noisy samples become larger, the uncertainty distribution of data changes very slightly. This demonstrates that although the noise may significantly pollute data, the neural network is still able to identify the underlying pattern even under large ratio of noisy data. Specifically, the noise inherent in data increases the difficulty to reconstruction thus produces corresponding larger uncertainty. Accordingly, the uncertainty estimated by each R-Net (for each view) reasonably guides the integration of multi-view data.
	
\begin{figure}[htp]
		\centering
		\subfigure[UCI-MF.]{
			\includegraphics[width=4.1cm]{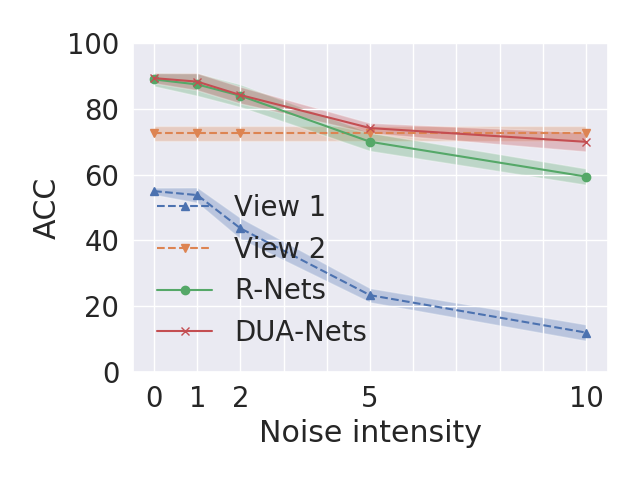}
		}\hspace{-10mm}
		\quad
		\subfigure[CUB.]{
			\includegraphics[width=4.1cm]{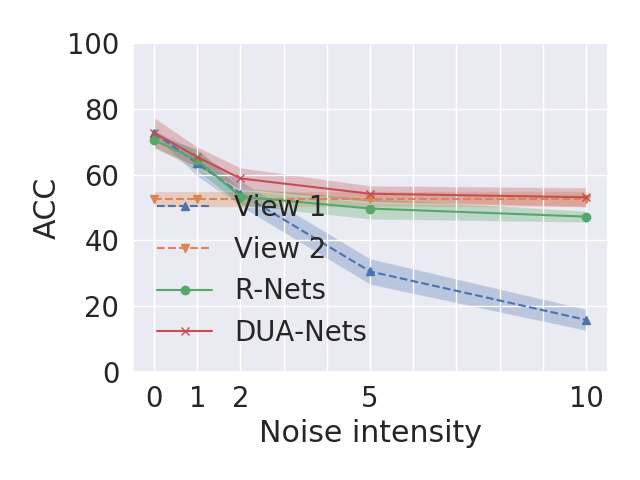}
		}
		\caption{Uncertainty in improving model performance. With the increasing of noise intensity, DUA-Nets can achieve a robust performance.}
		\label{fig:Noise_robust}
	\end{figure}

	\textbf{Uncertainty in improving model performance.}
	In Table~\ref{tab:clustering} and Table~\ref{tab:classification}, we show that the DUA-Nets are superior to R-Nets without uncertainty, which means that the learned uncertainty is beneficial for the representation learning. Here we conduct experiments to further verify the effect of uncertainty. We use UCI-MF dataset (with the first two views) and CUB dataset to conduct clustering task. In the experiment, view 1 of each dataset is selected to add Gaussian noise to half of the samples. As shown in Fig.~\ref{fig:Noise_robust}, with the increasing of the noise intensity, clustering performance on noisy view data decreases rapidly. However, the performance of DUA-Nets is quite stable. With the help of uncertainty, DUA-Nets are more robust to noise compared to R-Nets without uncertainty, which demonstrates that uncertainty can alleviate the influence of noisy observations.
	
	\subsection{Parameter Selection and Convergence}
\begin{figure}[]
		\centering
		\subfigure[CUB (original).]{
			\includegraphics[width=4.2cm]{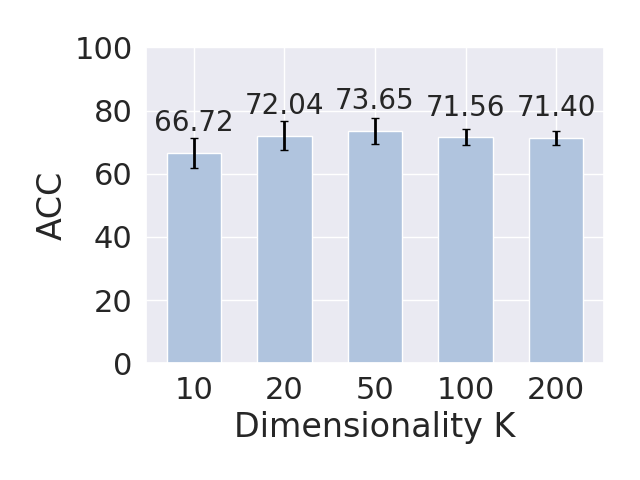}
		}\hspace{-5mm}
		\subfigure[CUB (noise intensity=1) .]{
			\includegraphics[width=4.2cm]{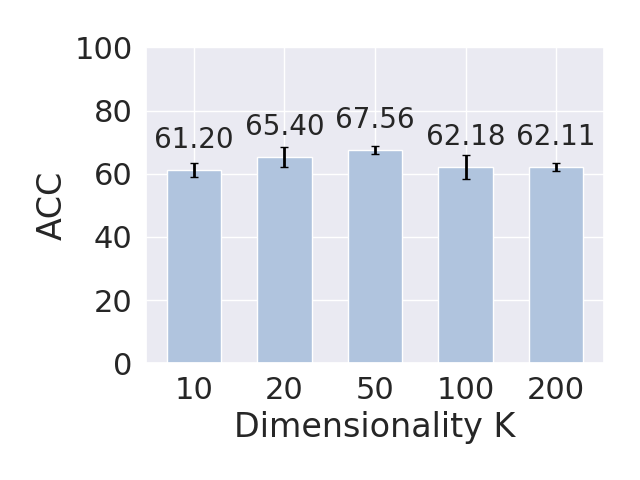}
		}
		\caption{Parameter tuning.  The performance of DUA-Nets with different dimensions for latent representation. }
		\label{fig:dimension}
	\end{figure}

\begin{figure}[]
		\centering
		\includegraphics[width=4.5cm]{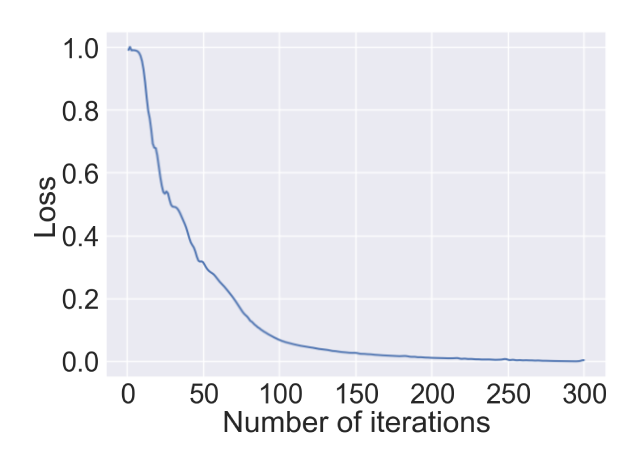}
		\caption{Convergence curve on CUB dataset (where loss values are normalized to range [0, 1]). The proposed method converges quickly within a small number of iterations.}
		\label{fig:loss}
	\end{figure}

	There is no explicit hyperparameter in our model, however, the dimension of latent representation needs to be specified in advance. In the experiments, we choose different dimensions of latent representation $\mathbf{h}_i$ to investigate its effect. We conduct clustering task on original CUB dataset as well as noisy CUB dataset (half of view 1 polluted by Gaussian noise). The dimensions are selected from [10, 20, 50, 100, 200]. As shown in Fig.~\ref{fig:dimension}, the best performance is obtained when the dimension is set to 50. As the dimension decreases too much, the latent representation may not have enough capacity to encode information from all views, which leads to a clear performance decline. Too larger dimensions also produce lower performance, where high-dimensional representation tends to overfit, and may contain possible noise in the final representation. Fig.~\ref{fig:loss} demonstrates the convergence of proposed method. Typically, the optimization process is basically stable, where the loss decreases quickly and converges within a number of iterations.
	
	\section{Conclusions}
	In this work, we propose a novel multi-view representation learning method incorporating data uncertainty. Our model considers the noise inherent in observations and weighs different views of different samples dynamically. The estimated uncertainty provides guidance for multi-view integration, which leads to more robust and interpretable representations. Extensive experiments show the effectiveness and superiority of our model compared to deterministic methods. We further provide insights of the estimated uncertainty by qualitative analysis. In the future, we will focus on more theoretical results to explain and support the improvement.
	\section{Acknowledgments}
	This work was supported in part by National Natural Science Foundation of China (No. 61976151, No. 61732011 and No. 61876127), the Natural Science Foundation of Tianjin of China (No. 19JCYBJC15200).
	\bibliographystyle{aaai}
	\bibliography{egbib}
	
\end{document}